# Graph Fuzzy System: Concepts, Models and Algorithms

Fuping Hu , Zhaohong Deng, *Senior Member*, *IEEE*, Zhenping Xie, Kup-sze Choi, Shitong Wang

*Abstract*—Fuzzy systems (FSs) have enjoyed wide applications in various fields, including pattern recognition, intelligent control, data mining and bioinformatics, which is attributed to the strong interpretation and learning ability. In traditional application scenarios, FSs are mainly applied to model Euclidean space data and cannot be used to handle graph data of non-Euclidean structure in nature, such as social networks and traffic route maps. Therefore, development of FS modeling method that is suitable for graph data and can retain the advantages of traditional FSs is an important research. To meet this challenge, a novel FS for graph data modeling called Graph Fuzzy System (GFS) is proposed in this paper, where the concepts, modeling framework and construction algorithms are systematically developed. First, GFS related concepts, including graph fuzzy rule base, graph fuzzy sets and graph consequent processing unit (GCPU), are defined. A GFS modeling framework is then constructed and the antecedents and consequents of the GFS are presented and analyzed. Finally, a learning framework of GFS is proposed, in which a kernel K-prototype graph clustering (K2PGC) is proposed to develop the construction algorithm for the GFS antecedent generation, and then based on graph neural network (GNNs), consequent parameters learning algorithm is proposed for GFS. Specifically, three different versions of the GFS implementation algorithm are developed for comprehensive evaluations with experiments on various benchmark graph classification datasets. The results demonstrate that the proposed GFS inherits the advantages of both existing mainstream GNNs methods and conventional FSs methods while achieving better performance than the counterparts.

*Index Term*s - Graph fuzzy system, graph fuzzy rule base, kernel K-prototype graph clustering (K2PGC), graph consequent processing unit (GCPU), graph classification.

## I. INTRODUCTION

Fuzzy systems (FSs) [1] are a kind of intelligent models based on fuzzy sets [2] and fuzzy logic [3]. They are important tools for dealing with uncertain knowledge representation and inference decision-making. FSs have been successfully applied in many fields, such as pattern recognition [4], intelligent control [5], data mining [6] and bioinformatics [7]. Compared with traditional intelligent models, FSs can better describe uncertainty and fuzziness that are common in the real world, and yield systems of stronger robustness [8]. FSs also have good interpretability because the systems can be expressed by IF-THEN rules based on language term description [9]. Besides, FSs have the data-driven learning capability similar to that of neural networks, which enabling the systems to learn a large amount of knowledge and laws from the data, and express them in the form of human interpretable rules [10].

Despite the excellent performance in processing images, texts and time series data, the existing FSs are limited to applications involving Euclidean spatial data. In recent years, graph data, a type of non-European spatial data, are gaining the attention of many researchers because of its ubiquitous existence in real life. Graph data can naturally represent the structure of data in the macro world or micro world, such as social networks, traffic routes and molecular structures. However, existing FSs cannot model graph data effectively and it is very meaningful to study novel fuzzy modeling techniques that are suitable for graph data while retaining the advantages of traditional fuzzy systems.

In graph data modeling research, the existing techniques can be divided into three main categories: graph kernel [11], graph embedding [12] and graph neural networks [13, 14]. Graph kernel is a traditional graph data modeling method. It generates the graph kernel matrix by using a kernel method suitable for graph data, and then transfers the graph kernel matrix to the downstream task execution module. Graph embedding is a common method for dealing with graph data mining problems, which converts graph data into low-dimensional spatial data while preserving the structural information and attribute information of the graph to the greatest extent. Graph neural networks (GNNs) are the mainstream method of graph learning in recent years. Classical GNNs are divided into two types: spectral-domain based GNNs and spatial-domain based GNNs [15]. The representative model of the former is Graph Convolutional Neural Network (GCN) [16] whereas the representative models of the latter are Graph Attention Network (GAT) [17] and GraphSAGE [18]. The commonly used mechanisms of GNNs include message propagation mechanism, attention mechanism, and neighborhood sampling mechanism. Although these modeling techniques have made significant advances in graph data modeling, there are still important challenges to tackle, which include representation of

This work was supported in part by the National key R & D plan under Grant (2022YFE0112400), the Chinese Association for Artificial Intelligence (CAAI)-Huawei MindSpore Open Fund under Grant CAAIXSLJJ-2021-011A, the NSFC under Grant 62176105, the NSFC under Grant 62272201, the Six Talent Peaks Project in Jiangsu Province under Grant XYDXX-056, the Hong Kong Research Grants Council (PolyU 152006/19E), the Project of Strategic Importance of the Hong Kong Polytechnic University (1-ZE1V) and the Postgraduate Research & Practice innovation Program of Jiangsu Province under Grant KYCX22_2314.

(Corresponding author: Zhaohong Deng).

F. P. Hu, Z. H. Deng, Z. P. Xie, and S. T. Wang are with the School of Artificial Intelligence and Computer Science, Jiangnan University and Jiangsu Key Laboratory of Media Design and Software Technology, Wuxi 214122, China (e-mail: hfping@stu.jiangnan.edu.cn; dengzhaohong@jiangnan.edu.cn; xiezp@jiangnan.edu.cn; wxwangst@aliyun.com).

K. S. Choi is with The Centre for Smart Health, the Hong Kong Polytechnic University, Hong Kong (e-mail: thomasks.choi@polyu.edu.hk).

uncertain knowledge in graph data, adaptation of uncertain graph data models, simulation of human inference ability and comprehensive multi-perspective decision making.

This paper proposes the Graph Fuzzy System (GFS) to deal with the challenges based on the characteristics of graph data and the relevant theories of traditional FSs. The GFS focus on the following research questions: (1) how to define the graph data fuzzy sets and assess the similarity of these graph data given the complex diversity of graph data, where different graphs contain different nodes and edges; (2) how to construct a fuzzy rule structure suitable for graph data modeling in order to achieve effective fuzzy inference and decision making; (3) how to effectively train GFS based on available graph data and utilize it for prediction of new samples. The main contributions are summarized as follows:

(1) GFS extends classical FSs to the fields of graph data learning by defining the key concepts of graph fuzzy sets, graph fuzzy rules and graph processing operators of the GFS.

(2) The GFS model architecture is proposed where the generation strategies, decision functions, training strategies and the corresponding neural network structure expressions are analyzed.

(3) The learning framework and algorithms of GFS are proposed. In particular, the kernel K-prototype graph clustering (K2PGC) algorithm is developed to generate the antecedents of the GFS. Furthermore, a GFS consequent generation method based on collaborative learning of the graph consequent processing unit (GCPU) is also proposed.

(4) The proposed GFS is thoroughly evaluated with extensive experiments, where three specific versions of GFS model are implemented and compared with classical data-driven TSK fuzzy systems and mainstream GNN models. Experiments on various different graph classification datasets demonstrate that the outstanding graph classification performance of GFS.

The remainder of the paper is organized as follows. Section II discusses the related work. The concepts, architecture and the construction methods of the GFS is proposed in Section III. Extensive experiments and the results are presented in Section IV to demonstrate the effectiveness of the proposed GFS. Finally, Section V draws the conclusions and suggests several future research directions. For clarity and easy reference, the abbreviations used in this paper are listed in Table I.

TABLE I ABBREVIATIONS USED IN THIS PAPER

| Abbreviations | Details |
|---|---|
| FS | Fuzzy system |
| ML-FS | Mamdani-Larsen fuzzy system |
| TSK-FS | Takagi-Sugeno-Kang fuzzy system |
| GFS | Graph fuzzy system |
| K2PGC | Kernel K-prototype graph clustering |
| GNN | Graph neural networks |
| GCN | Graph convolutional networks |
| GAT | Graph attention networks |
| GraphSAGE | Graph sample and aggregate |
| MLP | Multilayer perceptron |

## II. RELATED WORKS

In this section, we briefly discuss two related topics of the proposed GFS, i.e., graph data modeling and fuzzy systems.

### A. Graph Data Modeling

*1) Graph Data*

Denote graph data $G = (V, E)$, shown schematically in Fig. 1, where $V$ represents the vertex or node set in the graph, $E$ denotes the edge set, and $v_i \in V$ represents a node, $i = 1,2, \ldots, n$. Denote the edge between $v_i$ and $v_j$ by $e_{ij} = (v_i, v_j) \in E$. The neighborhood set of a node $v$ is defined as $N(v) = \{u \in V | (v,u) \in E\}$. Denote the adjacency matrix by $A$, which is an $n \times n$ real symmetric matrix; if $e_{ij} \in E$ then $A_{ij} = 1$, otherwise $A_{ij} = 0$. In this paper, the graph data that we mainly focus on are undirected graphs with node attributes. Here, $X$ denotes the feature matrix of all nodes in a graph, $X = [x_{v_1}; x_{v_2}; \ldots; x_{v_n}] \in \mathbf{R}^{n \times d}$, where $x_{v_i} \in \mathbf{R}^{1 \times d}$ is the feature vector of the node $v_i$. In addition, we use the notation $\mathcal{G}$ to denote a collection of multiple graphs.

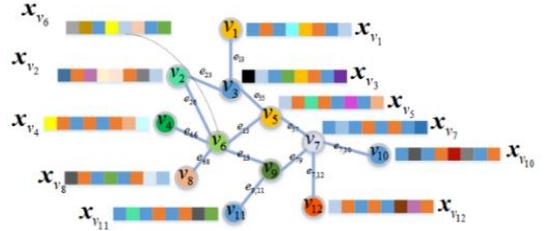

Fig.1. An example of graph data.

*2) Graph Kernel*

Graph kernel [11] is a kernel method to measure the similarity between graphs based on the inner product in the Hilbert space. The method can directly define the graph similarity measure on the graph structure. The graph kernel is a symmetric positive semi-definite function on the graph dataset $\mathcal{G}$. Once the function $k: \mathcal{G} \times \mathcal{G} \to \mathbb{R}$ is defined on the set $\mathcal{G}$, the mapping $\phi: \mathcal{G} \to \mathcal{H}$ to the Hilbert space $\mathcal{H}$ can be obtained such that

$$k(G_i, G_j) = \langle \phi(G_i), \phi(G_j) \rangle_{\mathcal{H}} \tag{1}$$

where $G_i, G_j \in \mathcal{G}$ and $\langle \cdot, \cdot \rangle_{\mathcal{H}}$ is the inner product in the Hilbert space $\mathcal{H}$.

*3) Graph Convolutional Neural Network*

Graph Convolutional Neural Network (GCN) [16] is a graph neural network model that incorporates generalized spectral features into convolutional layers to represent graph data. It is a neural network model constructed by stacking multiple graph convolutional layers. A single convolutional layer can be expressed as:

$$\boldsymbol{H} = \sigma(\widetilde{\boldsymbol{D}}^{-\frac{1}{2}} \widetilde{\boldsymbol{A}} \widetilde{\boldsymbol{D}}^{-\frac{1}{2}} \boldsymbol{X} \boldsymbol{W}) \tag{2}$$

where $\widetilde{\boldsymbol{A}} = \boldsymbol{A} + \boldsymbol{I}$ is the adjacency matrix with self-connection, $\boldsymbol{A}$ is the adjacency matrix, $\boldsymbol{I}$ is the identity matrix, $\widetilde{\boldsymbol{D}}$ is the degree matrix, $\widetilde{\boldsymbol{D}}_{ii} = \sum_{j=1}^{N_v} \widetilde{\boldsymbol{A}}_{ij}$, $\boldsymbol{X} \in \mathbb{R}^{N \times F}$ is the input feature matrix, $\boldsymbol{W} \in \mathbb{R}^{F \times F'}$ is the learnable weight, $\boldsymbol{H} \in \mathbb{R}^{N \times F'}$ is the convolved feature matrix, i.e., the output feature matrix, $F$ and $F'$ are the dimensions of the input feature matrix and the output feature matrix, respectively, and $\sigma(\cdot)$ is the nonlinear activation function, where the rectified linear unit (ReLU) function is generally used.

*4) Graph Attention Network*

Graph Attention Network (GAT) [17] is a classical attention-based graph neural network model that computes attention on the input graph data. When a node in the graph performs feature updates, it is necessary to perform attention calculation on its adjacent nodes and assign different attention coefficients to each adjacent node. The purpose of the mechanism is to highlight the influence of the more important neighbor nodes and weaken the influence of the less important ones during the process of node feature update. The feature update of each node in the graph can be expressed as:

$$\boldsymbol{H}_i^{l+1} = \sigma(\sum_{j \in \mathcal{N}_i} \alpha_{ij}^l \boldsymbol{H}_j^l \boldsymbol{W}^l) \qquad (3)$$

where $\mathcal{N}_i$ is the neighborhood set of node $v_i$, $\alpha_{ij}^l$ is the attention score of the neighbor node $v_j$ to node $v_i$ in the *l*-th layer GAT network ($l \geq 0$); $\boldsymbol{H}_j^l$ denotes the feature information of the neighborhood of node $v_i$ in the *l*-th layer; $\boldsymbol{W}^l$ is the learnable weight matrix of the *l*-th layer; and $\sigma(\cdot)$ is the nonlinear activation function. The common definition of $\alpha_{ij}^l$ is:

$$\alpha_{ij}^l = \frac{\exp(\text{LeakyReLU}(\mathcal{F}(\boldsymbol{H}_i^l \boldsymbol{W}^l, \boldsymbol{H}_j^l \boldsymbol{W}^l)))}{\sum_{k \in \mathcal{N}_i} \exp(\text{LeakyReLU}(\mathcal{F}(\boldsymbol{H}_i^l \boldsymbol{W}^l, \boldsymbol{H}_k^l \boldsymbol{W}^l)))} \qquad (4)$$

where $\mathcal{F}(\cdot,\cdot)$ is a mapping function that needs to be learned, such as the mapping function corresponding to a multilayer perceptron (MLP), and LeakyReLU is the activation function used to calculate attention.

*5) GraphSAGE*

GraphSAGE [18] is a general inductive graph representation learning framework that transforms graph data with unknown global structure information to node-level feature representation. GraphSAGE modifies GCN from two aspects. On the one hand, GraphSAGE samples neighbor nodes to transform GCN from a full-graph training method to a node-centered mini-batch training method. On the other hand, GraphSAGE extends the operation of aggregating neighbors and replaces graph convolution operations by several methods, e.g., GCN aggregation, mean aggregation, LSTM [19] aggregation and pooling aggregation.

*6) Graph Readout*

In graph classification datasets, each graph data has its unique structure, node features and edge features. In order to complete the label prediction of a single graph data, it is usually necessary to aggregate and summarize as much information as possible from a single graph data, and this type of operation is commonly called 'Readout' [20]. Common aggregation methods include summing, averaging, and maximizing or minimizing all node or edge features. For example, given a graph $G$, the aggregating readout operation that sums all its node features is:

$$\boldsymbol{H}_G = \sum_{v \in V} \boldsymbol{H}_v \qquad (5)$$

where $\boldsymbol{H}_G$ is the feature representation of graph $G$, $V$ is the set of nodes in graph $G$, $\boldsymbol{H}_v$ is the feature representation of node $v$. After obtaining $\boldsymbol{H}_G$, it can be passed to downstream modules for further processing.

*7) Graph-level Learning*

Graph classification is a classical graph-level learning task [21]. Graph-level learning requires the global information of the graph, which includes the structure information of the graph and the attribute information of each node. Assuming that each graph data in the graph set $\mathcal{G}$ is associated with a certain label $y \in \mathcal{Y}$, where $\mathcal{Y}$ is a predefined label set, the goal of the graph classification task is to learn a mapping function $f: \mathcal{G} \to \mathcal{Y}$, such that the function can accurately predict the class of the unknown graph. Many real-world problems can be formulated as graph classification problems, such as biochemical structure classification [22], social network classification [23] and user influence prediction [24].

*B. Fuzzy Systems*

A classical rule-based fuzzy system consists of four fundamental parts: fuzzifier, rule base, inference engine and defuzzifier [1]. There are currently two major FSs: Takagi-Sugeno-Kang fuzzy system (TSK-FS) [25, 26] and Mamdani-Larsen fuzzy system (ML-FS) [27, 28]. TSK-FS has been widely studied in recent years due to its powerful data-driven learnability [29].

For a TSK-FS with *d*-dimensional input and 1-dimensional output, the most commonly used fuzzy-inference rules can be expressed as follows:

IF $x_1$ is $A_1^k \wedge x_2$ is $A_2^k \wedge \cdots \wedge x_d$ is $A_d^k$,

THEN $f^k(\boldsymbol{x}) = p_0^k + p_1^k x_1 + p_2^k x_2 + \cdots p_d^k x_d$ (6)

where $k = 1,2,\ldots,K$, $k$ is the sequence number of fuzzy rules, and $K$ is the total number of fuzzy rules in the rule base; $\boldsymbol{x} = [x_1, x_2, \ldots, x_d]^\text{T}$ is the *d*-dimensional input vector; $A_i^k (i = 1,2,\ldots,d)$ denotes the fuzzy subset subscribed by the input variable $x_i$ for the *k*-th rule in the antecedents; $\wedge$ is a fuzzy conjunction operator; $f^k(\boldsymbol{x})$ is the output of the *k*-th rule; and $p_j^k (j = 1,2,\ldots,d)$ is the real-valued parameters of the rule consequent.

Attributed to outstanding data-driven learnability and interpretability, FSs have been successfully applied in many fields, e.g., intelligent control, pattern recognition, data mining and image processing. However, existing FSs are limited to applications involving Euclidean spatial data, which are not suitable for non-Euclidean spatial data such as graph data that are of great research interest in recent years.

III. GRAPH FUZZY SYSTEM

In this section, we propose a novel fuzzy system for graph data processing - Graph Fuzzy System. The concepts and models of GFS, the construction methods, and the relationship between GFS and existing methods are discussed.

*A. Concepts and Models*

*1) Rule Base*

Classical FSs construct a fuzzy rule base according to the fuzzy sets and fuzzy logic. Based on these rules, the inference from the input to the output can be achieved using fuzzy inference operations. Referring to the rule base form of classical TSK-FS, the graph fuzzy-inference rule base of GFS can be defined as follows.

**Definition 1 (GFS rule base):** The GFS rule base contains *K* rules in the following format:

$$\text{IF } G \text{ is } GF^1(\;), \quad \text{THEN } f^1(\;) = GCPU^1(\;),$$
$$\text{IF } G \text{ is } GF^2(\;), \quad \text{THEN } f^2(\;) = GCPU^2(\;),$$
$$\ldots$$
$$\text{IF } G \text{ is } GF^k(\;), \quad \text{THEN } f^k(\;) = GCPU^k(\;),$$
$$\ldots$$
$$\text{IF } G \text{ is } GF^K(\;), \quad \text{THEN } f^K(\;) = GCPU^K(\;).$$
$$k = 1,2,\ldots,K. \qquad (7)$$

In the GFS rule base defined in (7),

(i) the symbol denotes the input graph data;

(ii) $GF^k(\;)$ denotes the graph fuzzy set corresponding to the antecedent of the $k$-th fuzzy rule, which denotes a fuzzy set of graphs, and denotes the prototype graph of the fuzzy set, that is, the graph data sample that can best represent the fuzzy set;

(iii) $GCPU^k(\;)$ denotes the graph consequent processing unit (GCPU) corresponding to the $k$-th fuzzy rule consequent.

If $G$ is the input graph data of GFS, i.e., , and $G_{pt}^k$ is the prototype graph associated with the fuzzy set corresponding to the antecedent of the $k$-th fuzzy rule, i.e., , then the $k$-th rule in (7) can be simplified to the following general form:

$$\text{IF } G \text{ is } GF^k(G_{pt}^k), \text{ THEN } f^k(G) = GCPU^k(G) \qquad (8)$$

*2) Antecedents of GFS and Its Generation Strategy*

In GFS, each rule antecedent contains a graph fuzzy set, that is, a fuzzy set of graph data samples. The degree that an element belongs to the fuzzy set is determined by the membership function (MFs) of the fuzzy set. In order to conveniently describe the antecedent in GFS for the graph fuzzy set $GF^k(G_{pt}^k)$, i.e., IF $G$ is $GF^k(G_{pt}^k)$, we give the following definition.

**Definition 2 (Membership function of graph fuzzy set):** For the graph fuzzy set $GF^k(G_{pt}^k)$ that corresponds to each rule antecedent in GFS, its membership function can be defined as:
$$\mu_{GF^k}(G) = GraphSim(G, G_{pt}^k) \qquad (9)$$
where $k = 1,2,\ldots,K$ is the sequence number of fuzzy rules, $K$ is the total number of fuzzy rules; $G_{pt}^k$ is the prototype graph corresponding to the graph fuzzy set in the $k$-th rule, $G$ is an arbitrary graph to be evaluated, and the operator $GraphSim$ is a function used to calculate the similarity between $G_{pt}^k$ and $G$. The membership degree $\mu_{GF^k}(G)$ to the graph fuzzy set $GF^k(G_{pt}^k)$ in the rule antecedent can be calculated by $GraphSim(G, G_{pt}^k)$. In addition, $\mu_{GF^k}(G)$ can also be regarded as the firing strength corresponding to the $k$-th fuzzy rule for the input graph data $G$.

To construct the GFS antecedents, it is necessary to define the graph fuzzy set for each antecedent. In the regard, we can refer to the rule antecedent generation strategy of classical FSs, where the input space is usually partitioned appropriately to obtain the antecedents and the commonly used spatial partitioning technique is clustering. The classical clustering methods include k-means [30], fuzzy c-means(FCM) [31] and hierarchical clustering [32]. When clustering is used to divide the modeling data to generate the rules of FSs, each cluster of data obtained by clustering would give rise to a fuzzy rule, and the information of the clusters can be used to define the fuzzy set corresponding to the rule antecedent. Since the modeling data of GFS is graph data, classical clustering algorithms cannot be used to directly cluster graph data. Hence, it is necessary to develop an algorithm specifically for graph data clustering and to construct the GFS rule antecedents.

Based on the above analysis, a feasible GFS rule antecedent generation strategy is to generate the graph fuzzy sets of the rule antecedents by graph clustering. To construct $K$ fuzzy rules, we can cluster the available graph training data into $K$ clusters. We can then find the most representative graph data sample in each cluster as the prototype graph of the cluster. Here, the most representative graph data sample refers to the graph data sample with the largest sum of similarity with the other graph data samples in the same cluster. The prototype graph can be adopted as the parameter $G_{pt}^k$ of the graph fuzzy set $GF^k(G_{pt}^k)$ in (8) corresponding to the GFS rule antecedents. Hence, we can use the appropriate graph similarity measure function $GraphSim(G, G_{pt}^k)$ in (9) as the membership function of the graph fuzzy set. For example, graph kernel [33, 34], graph edit distance [35] and other techniques can be used to calculate the similarity between graphs, and then the membership function of the corresponding graph fuzzy set can be defined.

The current mainstream graph-clustering methods include MAPPR [36], MGAE [37], AGC [38], O2MAC [39] and DAEGC [40], which usually cluster nodes on a single graph and cannot be used to perform the graph-level clustering required for generating the fuzzy sets of the GFS rule antecedents. Here, we propose a framework that can perform the required graph-level clustering tasks by improving classical clustering methods with graph similarity measures. For example, hierarchical clustering can be combined with a graph similarity measure to realize a graph hierarchical clustering algorithm; k-means clustering can be combined with a graph similarity measure to develop a k-means clustering algorithm suitable for graph data sample set. These graph-similarity-measure based clustering methods can be readily used for graph data clustering to obtain the graph fuzzy sets to the GFS rule antecedents.

*3) Consequents of GFS and Its Analysis*

In addition to determining the antecedents of graph fuzzy rules, another important task in GFS is to learn and optimize the consequent parameters. As shown in (7), the consequent of the $k$-th rule of GFS is THEN $f^k\left(\tikz\right) = GCPU^k\left(\tikz\right)$. To describe the consequent more conveniently, we give the following definition.

**Definition 3 (Graph consequent processing unit (GCPU)):** For the consequent of the $k$-th rule of GFS, the graph consequent processing unit (GCPU) can be defined as:
$$f^k(G) = GCPU^k(G) \quad (10)$$
where $k = 1,2,\ldots,K$ is the sequence number of the graph fuzzy rules, $K$ is the total number of graph fuzzy rules and $G$ is the graph to be evaluated. GCPU is a modular unit that can directly deal with graph data. It can select different graph data learning components according to the needs of practical application scenarios, such as GCN, GAT and GraphSAGE. Once the GFS antecedent parameters and the specific model corresponding to the GCPU are determined, GFS consequent learning can be transformed into a collaborative optimization problem of multiple GCPU components parameters. Traditional models that can directly process graph data can be chosen as the model of GCPU. Among them, the classical GNN models [41, 42], which have good performance in graph data processing, are favorably used by GFS as the model of GCPU. While GNNs mainly deal with classification tasks at the node level, for graph-level classification, a graph-level feature generation module has to be added to the GCPU to aggregate node features into graph-level feature representations. Therefore, the consequent parameters learning of each GFS rule can be directly transformed into an optimization problem of jointly learning the parameters of traditional GNNs models and graph-level feature generation models. We describe the consequent parameters learning algorithm for GFS in detail in Section III-B-3).

*4) Decision-making of GFS*

The decision function of GFS can take different forms depending on the GCPU and the fuzzy inference operator used. Referring to the commonly used form of the decision function in the classical TSK-FS, we give the following decision function of GFS:

$$f_{GFS}(G) = \sum_{k=1}^{K} \tilde{\mu}_{\substack{k \\ GF}}(G) \cdot GCPU^k(G)$$
$$= \sum_{k=1}^{K} \tilde{\mu}_{\substack{k \\ GF}} \cdot Trans^k(Agg^k(G))$$
$$s.t. \sum_{k=1}^{K} \tilde{\mu}_{\substack{k \\ GF}} = 1 \quad (11)$$

where $\tilde{\mu}_{\substack{k \\ GF}} = \frac{\mu_{\substack{k \\ GF}}}{\sum_{k'=1}^{K} \mu_{\substack{k' \\ GF}}}$ is the normalized graph fuzzy membership of the input sample to the graph fuzzy set in the antecedent of the $k$-th fuzzy rule; $GCPU^k(G)$ is the consequent of GFS to process the graph $G$ in the $k$-th fuzzy rule; $Agg^k(G)$ is the output of the $k$-th rule after feature aggregation of the input graph $G$; $Trans^k(\cdot)$ is the feature transformation operation of the $k$-th rule, which includes the nonlinear activation function of the rule and the weight parameters to learn.

Among all the $K$ rules, each rule can use different GCPUs (e.g., different GNNs) to perform different $Agg$ and $Trans$ operations. It can be seen from (11) that feature aggregation and transformation of each rule is the core of the decision making of the GFS rule consequent. Node feature and structure feature of graphs are two important sources of information in the learning process. The node feature itself contains rich and complex information, and the feature representation of the node can be extracted by the $Agg$ operation. An important difference between graph data and Euclidean spatial data is the structure information of the graph, which can guide the feature learning of graph. Besides, it can be seen from (11) that the proposed GFS can be regarded as a comprehensive, dynamically weighted graph data processing model. For an arbitrary graph data input, all GCPUs corresponding to the graph fuzzy rules make decisions collaboratively and output the results of the comprehensive decision in the form of dynamic weighting with $\tilde{\mu}_{\substack{k \\ GF}}$ as the weight of the $k$-th fuzzy rule.

*5) Optimization of GFS*

The training of GFS is mainly the learning of GCPU parameters in the consequent of rules. For the prediction task of graph classification, the learning of the model-related parameters can be implemented by optimizing the following objective function:
$$\mathcal{L}_{GFS} = \mathcal{L}_{loss} + \alpha \mathcal{L}_{reg} \quad (12)$$
where $\mathcal{L}_{loss}$ is the loss function used to measure the error between the predicted output and the real output; $\mathcal{L}_{reg}$ is the regularization term used to improve the anti-disturbance ability of GFS; $\alpha > 0$ is the regularization parameter. By minimizing $\mathcal{L}_{GFS}$, GFS can optimize all consequent parameters in an end-to-end manner.

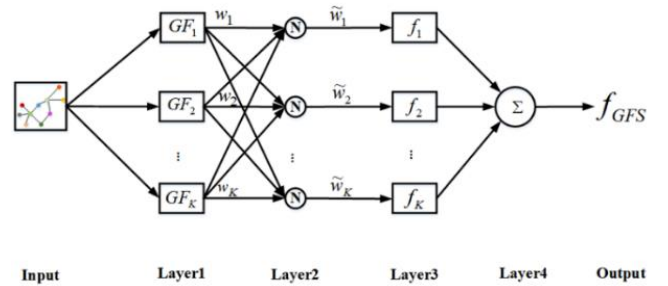

Fig. 2. The neural network structure of GFS.

*6) Neural Network Structure of GFS*

Like the classical FSs that can be equivalently represented as a special neural network, the proposed GFS can also be equivalently represented as a neural network whose structure is shown in Fig. 2. The neural network contains four layers which are described as follows.

**Layer 1 (Graph fuzzy rule antecedents processing layer):**
$$O_k^1 = w_k = \mu_{\underset{GF}{k}}(G), k = 1,2,\ldots,K \quad (13)$$

This layer is used to calculate the response of each rule antecedent to the input graph data, where $O_k^1$ is the output of the layer node; $w_k = O_k^1$ is also used to denote the firing strength of the $k$-th rule activated by the input graph data; $\mu_{\underset{GF}{k}}(\cdot)$ is the membership function of the graph fuzzy set corresponding to the antecedent of the $k$-th rule, and $K$ is the total number of graph fuzzy rules.

**Layer 2 (Normalization layer):**
$$O_k^2 = \widetilde{w}_k = \frac{w_k}{\sum_{k=1}^K w_k} \quad (14)$$

This layer normalizes the firing strength of the previous layer, $\widetilde{w}_k$ is the normalized firing strength, and $O_k^2$ is the output of this layer.

**Layer 3 (Consequent layer):**
$$O_k^3 = y_k = \widetilde{w}_k f_k \quad (15)$$

This layer executes the consequent processing of each rule, $O_k^3$ is the output of the node in this layer, and each node corresponds to a fuzzy rule; $f_k$ is the response of the GCPU of each rule, and $y_k$ is the final output of the $k$-th rule.

**Layer 4 (Summation layer):**
$$O^4 = \sum_{k=1}^K y_k \quad (16)$$

This layer is the comprehensive decision layer, and $O^4$ is the output of this layer, which is the integration of all rule outputs.

### B. Graph Fuzzy System Construction

In this section, we will study the construction algorithm of GFS. The framework of the proposed algorithm is first discussed, followed by the two main parts of the framework.

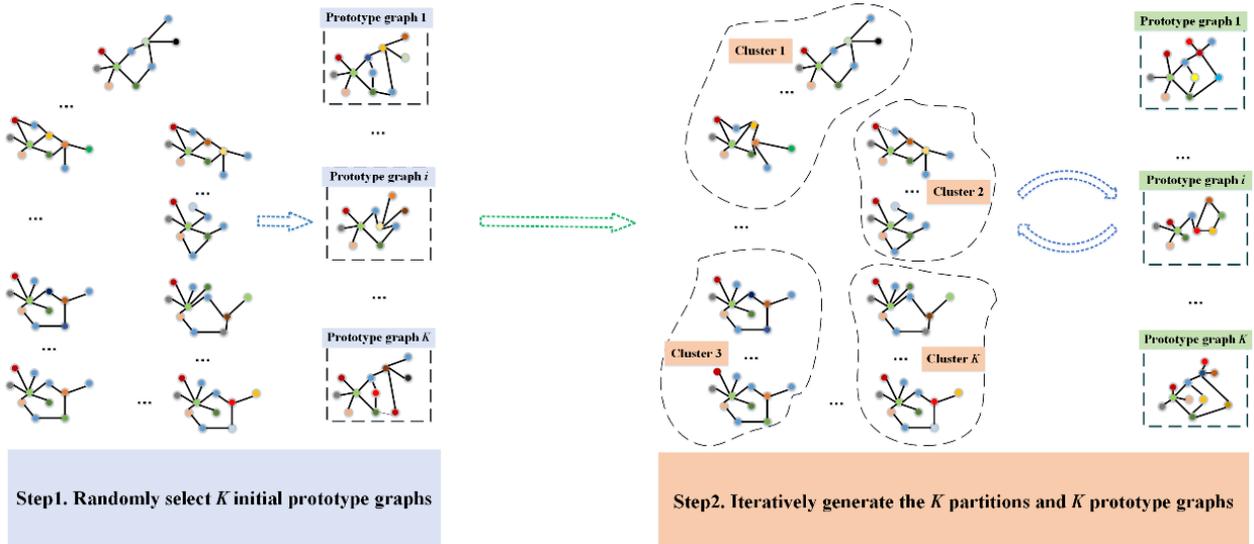

Fig. 3. The learning framework of GFS.

*1) Learning Framework*

The learning framework of GFS, as shown in Fig. 3, is proposed based on the GFS antecedent and consequent generation strategies described in Section III-A. The framework consists of two main parts. The first part concerns the generation of GFS rule antecedents based on kernel K-prototype graph clustering (K2PGC). It is to obtain representative graph data samples for each cluster in the graph data training set, thereby yielding prototype graphs through graph-level clustering. After obtaining the prototype graphs, the graph fuzzy set of the antecedents of the graph fuzzy rules can be generated using (8) and (9), and the antecedents of each rule can then be generated. In the second part, GNNs are used as the processing unit of the GFS consequent, and the objective function is constructed based on the cross-entropy loss and the L2-norm penalty term to optimize the parameters that need to be learned in the GCPU of the GFS consequents. These two parts are further discussed in the following two subsections.

Fig. 4. The framework of the Kernel K-Prototype Graph Clustering (K2PGC).

*2) Antecedent Generation based on K2PGC*

The framework of the K2PGC is shown in Fig. 4, which consists of two steps. In the Step 1 of Fig. 4, $K$ graphs are randomly selected from the input graph dataset as the initial prototype graphs. And in the Step 2 of Fig. 4, $K$ partitions and $K$ prototype graphs are iteratively generated using the classical graph similarity measurement technique. Furthermore, the Propagation Kernel (PK) [43] technique is used to measure the similarity between the graphs in our work. PK defines the kernel contribution for the $t$-th information propagation iteration, which is formulated as follows:

$$K\_Sim\left(G_t^{(i)}, G_t^{(j)}\right) = \sum_{u \in G_t^{(i)}} \sum_{v \in G_t^{(j)}} k(u,v) \quad (17)$$

where $K\_Sim\left(G_t^{(i)}, G_t^{(j)}\right)$ is the kernel similarity after the *t*-th information propagation iteration, $i, j = 1, 2, \ldots, N$; $u$ is the node in the $G^{(i)}$, $v$ is the node in the $G^{(j)}$, and $k(u,v)$ is an arbitrary node kernel determined by the node labels or node attributes. An important feature of PK is that the node kernel $k(u,v)$ is defined according to the probability distribution $p_{t,u}$ and $p_{t,v}$ corresponding to the node after the *t*-th information propagation iteration, and it is constantly updated during the propagation process. The node kernel between two graph nodes with labels and attributes is defined as:

$$k(u,v) = k_l(u,v) \cdot k_a(u,v) \quad (18)$$

where $k_l(u,v)$ is a kernel function related to the label information, and $k_a(u,v)$ is a kernel function related to the attribute information. If the label information does not exist, then $k(u,v) = k_a(u,v)$; if the attribute information does not exist, then $k(u,v) = k_l(u,v)$. The PK after $t_{MAX}$ iterations can be expressed as:

$$K\_Sim_{t_{MAX}}\left(G_t^{(i)}, G_t^{(j)}\right) = \sum_{t=1}^{t_{MAX}} K\_Sim\left(G_t^{(i)}, G_t^{(j)}\right) \quad (19)$$

with $i, j = 1, 2, \ldots, N$.

Based on the PK similarity measurement, each graph can be assigned to a cluster according to the nearest neighbor principle, that is, the similarity between each graph and the prototype graph of each cluster is first calculated, and the cluster to which the graph belongs is determined based on which prototype graph has the greatest similarity. This strategy can be expressed as follows:

$$\lambda_j = \arg\max_k K\_Sim_{t_{MAX}}\left(G_{pt}^k, G_j\right) \quad (20)$$

where $j = 1, 2, \ldots, N$, $\lambda_j$ is the cluster label of the *j*-th graph; $k = 1, 2, \ldots, K$, $G_{pt}^k$ is the prototype graph of the *k*-th cluster.

In order to effectively implement the K2PGC algorithm, the following prototype graph generation mechanism is developed. First, we update the prototype graph of each cluster $\{G_{pt}^1, G_{pt}^2, \ldots, G_{pt}^K\}$ according to the graph data contained in each current cluster. For each current cluster, each sample is taken as the center and the similarity between the sample and the other samples in the cluster is calculated. The sum of the similarity scores is then calculated. For each cluster, the sample with the highest similarity obtained by taking it as the center is the updated prototype graph of the cluster, i.e.,

$$C_k = \{G_j | \lambda_j = k\} \quad (21)$$

$$G_{pt}^k = \arg\max_{G_i \in C_k} \sum_{j=1}^{N_k} K\_Sim_{t_{MAX}}(G_i, G_j) \quad (22)$$

where $C_k$ is the sample set of the current *k*-th cluster, $k = 1, 2, \ldots, K$, $N_k$ is the number of samples in $C_k$, $t_{MAX}$ is the maximum number of iterations of the PK, and we set this value to 5. Here, we perform iterative learning of *K* cluster partitions and generation of *K* prototype graphs according to (21) and (22). To effectively divide the graph data into different clusters, K2PGC is essentially an optimization problem with the following objective function:

$$\max \mathcal{J}_{K2PGC}\left(G_{pt}^k, \lambda_i\right) = \sum_{k=1}^K \sum_{i=1}^N K\_Sim_{t_{MAX}}(G_i, G_{pt}^k) \quad (23)$$

where $K\_Sim_{t_{MAX}}(G_i, G_{pt}^k)$ denotes the similarity generated by PK between the *i*-th graph and the *k*-th prototype graph, $G_{pt}^k$ and $\lambda_i$ are the prototype graph of the *k*-th cluster and the cluster labels of the *i*-th graph respectively.

Based on the PK similarity measurement technique and prototype graphs generation strategy discussed above, the proposed K2PGC clustering algorithm is given in Algorithm 1.

---
**Algorithm 1**: K2PGC

**Input:** Graph dataset $\mathcal{G} = [G_1, G_2, \ldots, G_N]^T$, and number of clusters, i.e., *K*.

**Output:** Set of clusters $\mathcal{C} = \{C_1, C_2, \ldots, C_K\}$ and set of cluster prototype graphs $\{G_{pt}^1, G_{pt}^2, \ldots, G_{pt}^K\}$.

**Procedure:**
1: Initialize the cluster prototype graphs: randomly select *K* graphs from $\mathcal{G}$ as the initial prototype graphs of *K* clusters (analogous to random initiation of cluster centers in traditional *K* means clustering), expressed as $\{G_{pt}^1, G_{pt}^2, \ldots, G_{pt}^K\}$.
2: Update the clustering division according to (20) and (21).
3: Update the cluster prototype graphs according to (22).
4: Calculate the current value of objective function according to (23).
5: Repeat steps 2-4 until the objective functions value of multiple continuous iterations remain unchanged.

---

For a given graph data training set, the GFS antecedent generation process based on K2PGC is as follows. For the construction of *K* fuzzy rules, the graph data training set are clustered into *K* clusters, and then the prototype graphs are taken as the prototype graphs of the graph fuzzy set corresponding to the rule antecedent of the graph fuzzy system. Any suitable similarity function can be used to define the graph fuzzy membership function in (9). Accordingly, the antecedent part of the whole GFS can be determined.

*3) Consequent Generation of GFS based on GNNs*

The structure of the GCPU proposed for GFS is shown in Fig. 5. It consists of an input layer, 3-layers GNN, 1-layer graph summation readout, and a 3-layers MLP [44]. Among them, the 3-layers GNN is responsible for updating all node features in the graph data; the 1-layer graph summation readout operation is responsible for generating graph-level features representation; and the MLP is responsible for outputting graph classification prediction results. Based on the framework, the classical GCN, GAT and GraphSAGE in GNNs are used respectively to build three different versions of GCPU of the GFS rule consequent, denoted as GCPU-GCN, GCPU-GAT, and GCPU-GraphSAGE respectively. The only difference among them is the 3-layers GNN that updates the node features.

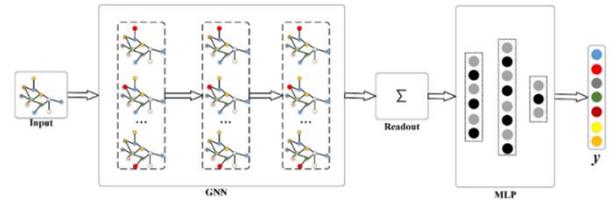

Fig. 5. The structure of GCPU.

*a) GCN-based GCPU*

The GCN-based GCPU is constructed using GCN as the GNN in the GCPU. The aggregation function in the GCPU is developed using a 3-layers GCN, which can be expressed as

$$Agg_{gcn}(A, X) = \sigma\left(\widetilde{D}^{-\frac{1}{2}}\widetilde{A}\widetilde{D}^{-\frac{1}{2}}\sigma\left(\widetilde{D}^{-\frac{1}{2}}\widetilde{A}\widetilde{D}^{-\frac{1}{2}}\sigma\left(\widetilde{D}^{-\frac{1}{2}}\widetilde{A}\widetilde{D}^{-\frac{1}{2}}XW^{(0)}\right)W^{(1)}\right)W^{(2)}\right) \quad (24)$$

where $\widetilde{A} = A + I$, $\widetilde{A}$ is the adjacency matrix with self-connection added, $A$ is the adjacency matrix, and $I$ is the identity matrix; $\widetilde{D}$ is the degree matrix, $\widetilde{D}_{ii} = \sum_{j=1}^{N_v} \widetilde{A}_{ij}$; $W^{(0)}$, $W^{(1)}$ and $W^{(2)}$ are the weight parameters that need to be learned for Layers 1-3 of the GCN, respectively. The final output of GCN-based GCPU can be expressed as follows:

$$f_{GFS-GCN}(G) = \sum_{k=1}^{K} \tilde{\mu}_{\underset{GF}{k}}(G) \cdot MLP^k(ReadOut_{sum}^k(Agg_{gcn}(A, X))) \quad (25)$$

where $ReadOut_{sum}^k$ is the graph summation readout operator of the $k$-th rule, which integrates the features of all nodes in the graph into a graph-level feature vector and $MLP^k$ is the MLP graph classification predictor of the $k$-th rule.

*b) GAT-based GCPU*

The GAT-based GCPU is constructed using the single-head GAT as the GNN in the GCPU. The aggregation function in the GCPU is developed using a 3-layers single-head GAT, which can be expressed as

$$Agg_{gat}(A, X) = GAT^2(GAT^1(GAT^0(A, X))) \quad (26)$$

where $GAT^0$, $GAT^1$ and $GAT^2$ are respectively the GAT network of the first, second and the third layer implemented by the methods in (3) and (4). The final output of the GAT-based GCPU is expressed as:

$$f_{GFS-GAT}(G) = \sum_{k=1}^{K} \tilde{\mu}_{\underset{GF}{k}}(G) \cdot MLP^k(ReadOut_{sum}^k(Agg_{gat}(A, X))) \quad (27)$$

where $ReadOut_{sum}^k$ is the graph summation readout operator of the $k$-th rule, which integrates the features of all nodes in the graph into a graph-level feature vector and $MLP^k$ is the MLP graph classification predictor of the $k$-th rule.

*c) GraphSAGE-based GCPU*

The GraphSAGE-based GCPU is constructed using GraphSAGE as the GNN in the GCPU. The aggregation function in the GCPU is developed using a 3-layers GraphSAGE, which can be expressed as

$$Agg_{gsage}(A, X) = GSAGE^2(GSAGE^1(GSAGE^0(A, X))) \quad (28)$$

where $GSAGE^0$, $GSAGE^1$ and $GSAGE^2$ are respectively the GraphSAGE network of the first, second and the third layer. The GCN method is used in the GraphSAGE for neighborhood feature aggregation, and the final output is given by

$$f_{GFS-GraphSAGE}(G) = \sum_{k=1}^{K} \tilde{\mu}_{\underset{GF}{k}}(G) \cdot MLP^k(ReadOut_{sum}^k(Agg_{gsage}(A, X))) \quad (29)$$

where $ReadOut_{sum}^k$ is the graph summation readout operator of the $k$-th rule, which integrates the features of all nodes in the graph into a graph-level feature vector and $MLP^k$ is the MLP graph classification predictor of the $k$-th rule.

*d) Model Optimization*

The objective function for graph classification tasks of GFS is developed to facilitate the learning and optimization of the GFS consequent parameters. The function can be expressed as

$$\min \mathcal{L}_{GFS}(\Theta) = -\sum_{i=1}^{N}\sum_{c=1}^{C} Y_{i,c} \log \hat{Y}_{i,c} + \alpha \|\Theta\|_2 \quad (30)$$

where $-\sum_{i=1}^{N}\sum_{c=1}^{C} Y_{i,c} \log \hat{Y}_{i,c}$ is the cross-entropy loss of graph classification, $N$ is the total number of graphs, $C$ is the number of categories, $Y \in \mathbb{R}^{N \times C}$ is the real label matrix of $N$ graphs with the one-hot encode, $Y_{i,c}$ denotes the real probability that the $i$-th graph belongs to $c$-th category, $\hat{Y} \in \mathbb{R}^{N \times C}$ is the predicted label matrix of $N$ graphs, $\hat{Y}_{i,c}$ denotes the predicted probability that the $i$-th graph belongs to $c$-th category, $\|\Theta\|_2$ is the L2 regularization term for the parameters of the GFS consequents, and $\alpha$ ($\alpha > 0$) is the trade-off parameter.

During the training process of GFS, the Adam optimization algorithm [45] is used to iteratively optimize the objective function. When the training set is large and cannot be put entirely in the GPU memory, as in the experiments to be presented, the mini-batch training strategy is adopted where the training data is divided into multiple batches for training. In addition to using the regularization term, the early stopping strategy is also used to prevent overfitting. In this study, the threshold for early stopping is set to 20, i.e., model training stops when the accuracy of the validation set does not increase for 20 consecutive rounds of iterations. At this point, the model parameters with the highest accuracy in the validation set during the training process are saved, which are then loaded for the testing.

*4) The GFS Algorithm*

Based on the discussions in Sections III.A, III-B-1), III-B-2) and III-B-3), the GFS construction algorithm for graph classification are given in Algorithm 2.

---

**Algorithm 2**: Construction of GFS

**Input**: Graph dataset: $\mathcal{G} = [G_1, G_2, ..., G_N]^T$;
Graph label: $Y = [y_1, y_2, ..., y_N]^T$;
Number of rules: K;
Batch size;
Epoch: maximal number of epochs;
Conditions of early stopping.

**Output**: Model parameters $\Theta$

**Procedure**:
*// Antecedent generation*
1: Run Algorithm 1 on the graph dataset to get K cluster prototype graphs.
2: According to the obtained cluster prototype graphs, generate the graph fuzzy membership function for the graph set in the antecedent of each fuzzy rules by (9) and (19).
3: Divide graph dataset $\mathcal{G}$ into a training set and a validation set, and then construct mini-batch data subsets from the training set according to the value of batch size.
*// Consequent generation*
4: **for** epoch=1 to Epoch **do**
5:   **for** each batch **do**
6:     Run Adam optimizer to minimize $\mathcal{L}_{GFS}$:
7:     $\Theta := \underset{\Theta}{\arg\min} \mathcal{L}_{GFS}$
8:     Calculate the training accuracy, validation accuracy, training loss, and validation loss for each batch.
9:   **end for**
10:   Calculate the average training accuracy, validation accuracy, training loss, and validation loss for each epoch according to the results obtained in Step 8 and save temporary optimal parameters $\Theta$.
11:   if early stopping conditions are satisfied
12:     Save parameters $\Theta$ with the optimal validation accuracy.
13:     break
14: **end for**
15: Output model parameters $\Theta$ with the optimal validation accuracy.

---

*5) Computational Time Complexity*

The computational time complexity of the GFS algorithm is analyzed in this section. The GFS algorithm consists of two main parts: 1) using the K2PGC clustering algorithm to obtain the antecedent parameters of the graph fuzzy rules; 2) learning the consequent parameters based on the cross-entropy loss function of GNNs and L2-norm penalty terms. For the first part, the computational complexity of the K2PGC algorithm is

$O(TNKd_{in})$, where $T$ is the number of iterations of the K2PGC algorithm, $N$ is the number of graph data, and $K$ is the total number of graph fuzzy rules. For the second part, the learning of consequent parameters involves joint optimization of GNNs and MLP. Here, we take GFS-GCN as an example. We first analyze the time complexity of the GCN employed in the GFS consequent. Let $n$ be the number of nodes in a single graph; $m$ be the number of edges in a single graph; $d_{in}$ be the dimension of node features; $d_h$ be the dimension of hidden layer features of the nodes; $L$ be the number of network layers of GNNs in the GCPU of the GFS rule consequents; $d_{MLP}$ be the feature dimension of the hidden layer nodes of the MLP used in GFS; $d_{out}$ be the feature dimension of the final output of GFS. The time complexity of GCN is $O(Lmd_h + Lnd_{in}d_h)$, and the time complexity of GFS-GCN consequent parameters learning is $O(K(Lmd_h + Lnd_{in}d_h + d_hd_{MLP}d_{out}))$. Therefore, the time complexity of the proposed GFS-GCN algorithm is $O(K(TNd_{in} + Lmd_h + Lnd_{in}d_h + d_hd_{MLP}d_{out}))$. Similarly, the time complexity of GFS-GAT and GFS-GraphSAGE are: $O(K(TNd_{in} + Lmd_h + Lnd_{in}d_h + d_hd_{MLP}d_{out}))$ and $O(K(TNd_{in} + r^L nd_{in}d_h + d_hd_{MLP}d_{out}))$, respectively, where $r$ is the number of sampled neighborhood nodes of each node in GraphSAGE.

### C. Relationship between GFS and Existing Methods

The relationship between the proposed GFS and classical FSs, as well as the relationship between GFS and ensemble learning, are analyzed in this section.

#### 1) Relationship Between GFS and Classical FSs

Classical fuzzy systems construct intelligent prediction models based on fuzzy sets and fuzzy rules. At present, the most widely used data-driven fuzzy system is TSK fuzzy system (TSK-FS). The proposed GFS belongs to this category of fuzzy systems, but it is significantly different from the classical data-driven fuzzy systems. Taking TSK-FS as an example, the common strategy of TSK-FS in fuzzy rules antecedents learning is to use the classical clustering algorithm to generate the fuzzy rules antecedents, while GFS needs to utilize the graph clustering algorithm according to the characteristics of graph data. In addition, the fuzzy rules consequents of TSK-FS usually adopt a linear regression model that cannot be used to process graph data directly and effectively, while the fuzzy rules consequent of GFS is a graph consequent processing unit (GCPU) which can make use of any model that has the ability to process graph data. Therefore, the proposed GFS extends the classical fuzzy system by developing novel methods for constructing antecedents and consequents that can directly process graph data.

#### 2) Relationship between GFS and Ensemble Learning

Ensemble learning methods can achieve better predictive performance than using any single learning method. The GFS proposed in this paper is similar to ensemble learning. Here, we discuss the relationship between GFS and the ensemble learning mechanism, as well as two of its representative methods, AdaBoost and AdaGCN.

##### a) Relevance to Ensemble Learning Mechanism

Ensemble learning [46], also known as multi-classifier systems or committee-based learning performs a learning task by constructing and combining multiple individual learners to achieve better generalization performance than a single learner. Ensemble learning can be categorized as homogeneous ensemble and heterogeneous ensemble, depending on whether the same kind of individual learners are used or not.

The GFS proposed in this paper is similar to an ensemble learning method in that it integrates multiple fuzzy rules through a fuzzy inference mechanism to perform comprehensive decision making. Each fuzzy rule is regarded as a sub-model and the whole GFS is thus an integrated model. Since the GCPU of GFS rules can flexibly adopt models of the same or different kinds according to the needs of application scenarios, GFS can realize both homogeneous integration and heterogeneous integration. The three GFSs discussed in Section III-B-3) - GFS-GCN, GFS-GAT, GFS-GraphSAGE belong to the category of homogeneous ensemble learning. Hence, GFS has the characteristics of ensemble learning and can be seen as a GNN ensemble learning method with fuzzy inference capability.

##### b) Relationship between GFS and AdaBoost

Classical AdaBoost [47] algorithm trains different classifiers (base classifiers, also known as weak classifiers) on the same dataset, and then aggregates the classifiers to construct a strong classifier. Freund et al. have demonstrated in [47] that AdaBoost outperforms its base classifier in most cases. In AdaBoost, different weak classifiers are trained on different weighted sets of samples. The classification difficulty of each sample determines the weight, whereas the classification difficulty is estimated by the output of the classifier in the previous step. By adjusting the sample weight and the weight of the base classifier, AdaBoost is finally combined individual classifiers to become a strong classifier.

In the GFS proposed in this paper, the GCPU constructed with different rules learn from the same graph dataset at the same time. Because of the different antecedents of the different rules, the same graph data have different effects on the consequents. That is, the consequents of different rules are trained under different sample weights of the same dataset, which is analogous to the different weights used in AdaBoost for training different base classifiers. In addition, the decision functions of GFS and AdaBoost both adopt the form of weighted comprehensive decision of multiple sub-models. The difference is that the weight of each sub-model in AdaBoost is fixed, while the weight of the sub-model (or rule) in GFS changes dynamically, where the weight (i.e., firing strength) of each sub-model can be adaptively changed according to the input of the decision function. Hence, GFS can be seen as more flexible ensemble method than AdaBoost, in addition to a novel FS inheriting the characteristics of the classical ones with direct graph data processing ability.

##### c) Relationship Between GFS and AdaGCN

AdaGCN [48] is a deep GCN [49] that incorporates AdaBoost into graph convolutional neural network. This method extracts feature knowledge from neighborhoods and combines the information through iterative updates of the node weights in a manner following AdaBoost. Unlike other GNNs models that directly stack graph convolutional layers, AdaGCN shares the same infrastructure among all the GCN layers and optimizes them in a loop. It can be seen that AdaGCN is still a cascaded GCN stack.

The GFS proposed in this paper performs parallel learning on graph data using the GCPUs in multiple graph fuzzy rules, and realizes the comprehensive decision making using fuzzy inference to integrate the multiple rules. As the GCPU in each rule is constructed with classical GCN, the GFS can be regarded as a parallel stack of different GCNs, which is similar to the structure adopted by broad learning that has attracted more attention in recent years.

In conclusion, the proposed GFS inherits the advantages of many classical methods, not only the fuzzy inference ability of classical FSs, but also the graph data learning ability of GNNs and the advantages of ensemble learning.

## IV. EXPERIMENTAL STUDIES

The proposed GFS is comprehensively evaluated with the three specific GFS versions, i.e., GFS-GCN, GFS-GAT and GFS-GraphSAGE, on eight graph classification datasets. The difference of three versions of GFS is that different consequent processing units, i.e., GCPU-GCN, GCPU-GAT, and GCPU-GraphSAGE are adopted, respectively.

### A. Datasets

Eight graph classification datasets used in the experiments are listed in Table II. The datasets can be downloaded from https://chrsmrrs.github.io/datasets/docs/datasets/. The samples of the datasets are all graph data with node feature information. In Table II, the headers "**# Graphs**", "**Avg. # Nodes**", "**Avg. # Edges**", "**# Node Features**" and "**# Categories**" are the total number of graphs, the average number of nodes, the average number of edges, the number of node features and the number of categories in the datasets respectively. The eight datasets are described in Part 1 of the *Supplementary Materials* section in detail.

TABLE II STATISTICS OF THE GRAPH CLASSIFICATION DATASETS

| Datasets | # Graphs | Avg. # Nodes | Avg. # Edges | # Node Features | # Categories |
|---|---|---|---|---|---|
| ENZYMES [50] | 600 | 32.63 | 62.14 | 18 | 6 |
| PROTEINS [51] | 1113 | 39.06 | 72.82 | 1 | 2 |
| PROTEINS_full [51] | 1113 | 39.06 | 72.82 | 29 | 2 |
| BZR [52] | 405 | 35.75 | 38.36 | 3 | 2 |
| DHFR [52] | 467 | 42.43 | 44.54 | 3 | 2 |
| COX2 [52] | 467 | 41.22 | 43.45 | 3 | 2 |
| Cuneiform [53] | 267 | 21.27 | 44.80 | 3 | 30 |
| AIDS [54] | 2000 | 15.69 | 16.20 | 4 | 2 |

### B. Experimental Settings

#### 1) Comparison Algorithms

Two groups of algorithms are compared with the proposed GFS to evaluate and verify the effectiveness. They are the classical FSs and classical GNNs, involving six TSK-FS based classification methods for the former and three representative GNNs for the latter. These two groups of comparison algorithms are described as follows.

*a) Six Classification Methods Based on TSK-FS*

**L2-TSK-FS** [55]: This is a TSK-FS modeling method based on L2-norm penalty.

**TSK-MBGD-UR-BN** [56]: This is a TSK-FS modeling method optimized using mini-batch stochastic gradient descent, uniform regularization and batch normalization.

**FCM-TSK-FS**: This is a conventional TSK-FS model which uses FCM [31] to generate the antecedent parameters of TSK-FS.

**EWFCM-TSK-FS [57]**: This method is a TSK-FS modeling method where the TSK-FS antecedent parameters are generated using an entropy-weighted FCM method.

**ESSC-TSK-FS** [58]: This is a TSK-FS modeling method that uses the enhanced soft subspace clustering method (ESSC) [59] to generate the antecedent parameters.

**SESSC-LSE** [58]: This modeling method initializes TSK-FS using a supervised soft subspace clustering method and optimizes the model using least squares evaluation.

*Remark 1*: The six FS modeling methods listed above can only deal with Euclidean spatial data. To enable these methods to process graph data, we perform a graph summation readout operation on the graph datasets to obtain a feature vector to represent the features of the entire graph.

*b) Three Classical GNN Methods*

**GCN** [16]: This is a classical GNN method which is commonly used as a benchmark in graph classification. It is adopted as the GNN in the GCPU to construct the proposed GFS-GCN.

**GAT** [17]: This is a representative GNN model that fosters the development of graph neural network by the application of the attention mechanism in graph neural network. It is adopted as the GNN in the GCPU to construct the proposed GFS-GAT.

**GraphSAGE** [18]: This is a representative spatial GNN model which updates the feature representation of nodes by sampling a fixed number of neighboring nodes. Although the original paper the GraphSAGE is not used for graph classification, it is often used as a benchmark method for graph classification tasks. It is adopted as the GNN in the GCPU to construct the proposed GFS-GraphSAGE.

*Remark 2*: To effectively perform graph classification with the three proposed GFSs, a layer of graph summation readout is used in each rule consequent after applying 3-layer GCN, 3-layer GAT and 3-layer GraphSAGE, respectively, to obtain the entire feature representation of the graph. With the graph-level feature representation, a 3-layer MLP is further used to produce the classification output of the rule.

#### 2) Hyperparameter settings

The hyperparameter settings of the algorithms in the experiments are as follows:

(1) For fuzzy systems, the numbers of fuzzy rules are searched from the set {2, 3, 4, 5, 6, 7, 8, 9, 10}. The other hyperparameters of L2-TSK-FS, TSK-MBGD-UR-BN, FCM-TSK-FS, EWFCM-TSK-FS, ESSC-TSK-FS and SESSC-LSE follow the default optimal settings reported in the respective original papers.

(2) The L2 regularization coefficients of GCN, GAT, GraphSAGE, GFS-GCN, GFS-GAT and GFS-GraphSAGE are searched from the set {1e-10, 1e-9, 1e-8, 1e-7, 1e-6, 1e-5, 1e-4, 1e-3, 1e-2, 1e-1, 1, 1e, 1e2, 1e3, 1e4, 1e5, 1e6}.

(3) The feature dimension of the nodes in each layer of the different GNNs involved in different algorithms is searched from the set {16, 32, 64, 128, 256}.

(4) The maximal number of epochs for GCN, GAT, GraphSAGE, GFS-GCN, GFS-GAT and GFS-GraphSAGE is set to 100, and the learning rate decays exponentially with the initial learning rate is set to 0.1 and the attenuation coefficient is set to 0.98.

Other settings of experiments are given in Part2 of the *Supplementary Materials* section

TABLE III CLASSIFICATION ACCURACY OF NINE ALGORITHMS ON EIGHT GRAPH DATASETS

| Methods | Datasets | | | | | | | | |
|---|---|---|---|---|---|---|---|---|---|
| | ENZYMES | PROTEINS | PROTEINS_full | BZR | DHFR | COX2 | Cuneiform | AIDS | Average |
| L2-TSK-FS | 29.50 (±3.56) | 59.56 (±4.03) | 59.57 (±2.22) | 78.77 (±6.78) | 61.11 (±5.20) | 78.17 (±5.62) | 6.00 (±5.38) | 88.75 (±1.72) | 57.68 (±4.31) |
| TSK-MBGD-UR-BN | 40.53 (±1.53) | 71.10 (±1.71) | 73.50 (±0.57) | 76.42 (±3.99) | 59.71 (±1.25) | 75.96 (±1.93) | 19.30 (±1.41) | 98.66 (±0.15) | 64.40 (±1.57) |
| FCM-TSK-FS | 48.83 (±3.71) | 71.97 (±2.39) | 74.49 (±1.59) | 79.01 (±0.78) | 60.45 (±1.47) | 78.59 (±1.42) | 18.71 (±3.06) | 98.24 (±0.52) | 66.29 (±1.68) |
| EWFCM-TSK-FS | 52.00 (±4.20) | 66.30 (±2.39) | 74.57 (±1.76) | 78.77 (±4.94) | 60.45 (±1.17) | 78.16 (±0.41) | 18.32 (±4.99) | 94.30 (±1.41) | 65.36 (±2.66) |
| ESSC-TSK-FS | 51.00 (±2.07) | 66.30 (±2.39) | 74.30 (±1.91) | 78.51 (±0.60) | 60.84 (±1.58) | 78.16 (±0.41) | 21.71 (±4.79) | 98.65 (±0.49) | 66.18 (±1.78) |
| SESSC-LSE | 42.50 (±4.86) | 70.98 (±2.21) | **75.11** (±**1.76**) | 78.77 (±0.49) | 61.24 (±0.39) | 78.16 (±0.41) | 14.59 (±5.59) | 92.60 (±1.59) | 64.24 (±2.16) |
| GFS-GCN(Ours) | 50.17 (±4.13) | 74.13 (±3.01) | 70.71 (±1.92) | 80.74 (±2.77) | 64.42 (±1.73) | 80.10 (±2.10) | 22.83 (±3.90) | **99.45** (±**0.37**) | 67.82 (±2.49) |
| GFS-GAT(Ours) | **54.67** (±**5.60**) | 73.95 (±1.50) | 72.96 (±2.61) | 82.22 (±1.67) | 68.79 (±1.38) | 79.46 (±3.03) | 22.47 (±4.83) | 99.00 (±0.52) | 69.19 (±2.64) |
| GFS-GraphSAGE (Ours) | 53.67 (±3.23) | **74.31** (±**2.83**) | 70.80 (±1.47) | **83.21** (±**3.28**) | **72.49** (±**4.19**) | **80.74** (±**3.51**) | **32.19** (±**6.16**) | 98.55 (±0.29) | **70.75** (±**3.12**) |

* The best results for a dataset are bold faced.

## C. Results and Discussion

### 1) Comparison with TSK-FSs

The classification performance of three GFS models proposed in this paper on the eight graph classification datasets are compared with that of the six TSK-FS based methods. The results are shown in Table III. For the five datasets, PROTEINS, BZR, DHFR, COX2, Cuneiform, and AIDS, the graph classification accuracy of the three GFS models is better than that of the six TSK-FS based methods. For the ENZYMES dataset, the accuracy of the proposed GFS-GAT and GFS-GraphSAGE is higher than the TSK-FS based methods. However, for the PROTEINS_full dataset, the accuracy of the three GFS models is slightly lower than that of the TSK-FS based methods. A possible reason is that the structure information in the graph is not helpful for classifying the whole graph, or even produces negative impact. Therefore, structure information does not necessarily improve the classification performance. Besides, it can be seen from Table III that the average classification accuracies of the three GFS models are higher than that of the TSK-FS based methods. Based on the above analysis, we conclude that the three GFS models, GFS-GCN, GFS-GAT and GFS-GraphSAGE, have a better graph classification performance than the traditional TSK-FSs.

### 2) Comparison with GNNs

The graph classification accuracy of the three GFS models is compared with that of the three traditional GNNs, i.e., GCN, GAT and GraphSAGE. The accuracies of GFS-GCN and GCN are compared in Table IV and Fig. S1(a) of the *Supplementary Materials* section, which show that GFS-GCN outperforms GCN in classifying six of the eight datasets (ENZYMES, PROTEINS, BZR, DHFR, COX2 and Cuneiform). The accuracies of GFS-GAT and GAT are compared in Table V and Fig. S1(b) of the *Supplementary Materials* section, which indicate that the performance of GFS-GAT is better than GAT in classifying seven datasets (ENZYMES, PROTEINS, PROTEINS_full, BZR, DHFR, COX2 and Cuneiform). The results in Table VI and Fig. S1(c) of the *Supplementary Materials* section show that GFS-GraphSAGE outperforms GraphSAGE in the classification of all the eight graph datasets.

It can be concluded from the results in Sections IV-C-1) and IV-C-2) that:

(1) The three GFS models show significantly better classification performance than the classical TSK-FSs. Therefore, GFS has shown a stronger learning ability for graph data.

(2) The three GFS models have shown promising performance, mainly attributed to their ability in expressing the uncertainty in graph data and that the multiple graph fuzzy rules can effectively cooperate during inference, performing human-like comprehensive decision making and ensemble learning.

TABLE IV CLASSIFICATION OF GFS-GCN AND GCN ON EIGHT GRAPH DATASETS

| Methods | Datasets | | | | | | | | |
|---|---|---|---|---|---|---|---|---|---|
| | ENZYMES | PROTEINS | PROTEINS_full | BZR | DHFR | COX2 | Cuneiform | AIDS | Average |
| GCN | 44.83 (±5.88) | 73.68 (±1.93) | **71.43** (±3.62) | 80.00 (±2.39) | 62.44 (±1.23) | 79.02 (±1.77) | 14.57 (±5.86) | 99.35 (±0.41) | 65.67 (±2.89) |
| GFS-GCN (Ours) | **50.17** (±4.13) | **74.13** (±3.01) | 70.71 (±1.92) | **80.74** (±2.77) | **64.42** ±1.73) | **80.10** (±2.10) | **22.83** (±3.90) | **99.45** (±0.37) | **67.82** (±2.89) |

* The best results for a dataset are bold faced.

TABLE V CLASSIFICATION ACCURACY OF GFS-GAT AND GAT ON EIGHT GRAPH DATASETS

| Methods | Datasets | | | | | | | | |
|---|---|---|---|---|---|---|---|---|---|
| | ENZYMES | PROTEINS | PROTEINS_full | BZR | DHFR | COX2 | Cuneiform | AIDS | Average |
| GAT | 51.17 (±4.61) | 72.78 (±3.23) | 71.25 (±4.88) | 80.99 (±2.42) | 67.07 (±2.52) | 79.25 (±3.87) | 18.74 (±7.44) | **99.40** (±0.25) | 67.58 (±3.65) |
| GFS-GAT (Ours) | **54.67** (±5.60) | **73.95** (±1.50) | **72.96** (±2.61) | **82.22** (±1.67) | **68.79** (±1.38) | **79.46** (±3.03) | **22.47** (±4.83) | 99.00 (±0.52) | **69.19** (±2.64) |

* The best results for a dataset are bold faced.

TABLE VI CLASSIFICATION ACCURACY OF GFS-GraphSAGE AND GraphSAGE ON EIGHT GRAPH DATASETS

| Methods | Datasets | | | | | | | | |
|---|---|---|---|---|---|---|---|---|---|
| | ENZYMES | PROTEINS | PROTEINS_full | BZR | DHFR | COX2 | Cuneiform | AIDS | Average |
| GraphSAGE | 48.00 (±6.94) | 72.41 (±3.24) | 70.26 (±2.59) | 82.47 (±9.2) | 71.04 (±2.64) | **80.74** (±3.75) | 27.34 (±1.41) | 98.50 (±1.08) | 68.85 (±3.86) |
| GFS-GraphSAGE (Ours) | **53.67** (±3.23) | **74.31** (±2.83) | **70.80** (±1.47) | **83.21** (±3.28) | **72.49** (±4.19) | 80.74 (±3.51) | **32.19** (±6.16) | **98.55** (±0.29) | **70.75** (±3.12) |

* The best results for a dataset are bold faced.

## D. Parameter Analysis

The hyperparameters of GFS-GCN, GFS-GAT and GFS-GraphSAGE, including the number of graph fuzzy rules, the feature dimension of hidden layer nodes in GNNs, and the regularization parameter in objective function are analyzed in this section.

### 1) Number of Graph Fuzzy Rules

The analysis in Section III-B-5) has indicated that the computational complexity of GFS increases linearly with the number of graph fuzzy rules. By further analyzing the influence of this parameter on classification performance, we can obtain a reference for choosing the appropriate number of rules to balance between computational complexity and classification accuracy in practical applications. Fig. S2 of the *Supplementary Materials* section describes the classification accuracy of three GFS models under different rule settings on the eight datasets.

For GFS-GCN, Fig. S2(a) of the *Supplementary Materials* section shows that high accuracy can be achieved with a few rules, e.g. two or three rules, for six of the eight datasets (AIDS, BZR, COX2, DHFR, PROTEINS and PROTEINS_full). For the ENZYMES dataset, the classification accuracy increases steadily with the increasing number of graph fuzzy rules. For the Cuneiform dataset, a higher classification accuracy can be obtained when the number of rules is eight.

Similar results are obtained for GFS-GAT, as shown in Fig. S2(b) of the *Supplementary Materials* section, where high accuracy can be achieved with 2 to 3 rules for six datasets (AIDS, BZR, COX2, DHFR, PROTEINS and PROTEINS_full). For the ENZYMES dataset, classification accuracy increases steadily with the number of rules increasing from 2 and 9, and then decreases when the number of rules exceeds 9. For the Cuneiform dataset, higher classification accuracy can be obtained when the number of rules is 6 or 7.

For, GFS-GraphSAGE, Fig. S2(c) of the *Supplementary Materials* section shows that high accuracy can be achieved for the PROTEINS_full, BZR, DHFR and AIDS datasets when the number of rules is 2. For COX2 and PROTEINS datasets, the number of rules for high classification accuracy is 3. For the ENZYMES dataset, high classification accuracy can be obtained when the number of rules is 6 or 7. For the Cuneiform dataset, the corresponding number of rules is 5 or 7.

The result of the above analysis shows that the three GFS models only need a small number of rules to obtain high classification accuracy on most of the datasets adopted, i.e., PROTEINS, PROTEINS_full, BZR, COX2, DHFR and AIDS. For the ENZYMES and Cuneiform datasets, a larger number of rules are required. While optimal results are obtained with different numbers of rules for different datasets, the variation in performance is not significant in a bigger range to set this parameter. For a balance between accuracy and complexity, a reasonable choice of the number of rules is between 2 and 7 for the GFS models.

### 2) Feature Dimension of Hidden Layer Nodes in GNNs

As the node feature learning modules in the consequents of GFS are GNNs, it is important to analyze the effect of the feature dimension of nodes in each layer of the GNN on the classification accuracy. Fig. S3(a)-(c) of the *Supplementary Materials* section show the effect of the feature dimension of hidden layer nodes in the GNNs for GFS-GCN, GFS-GAT and GFS-GraphSAGE on the eight datasets.

It can be seen from Fig. S5(a) of the *Supplementary Materials* section that for the seven datasets AIDS, PROTEINS, PROTEINS_full, BZR, DHFR, COX2 and Cuneiform, the classification accuracy of GFS-GCN decreases slightly with increasing feature dimension, indicating that better classification performance can be obtained with a lower dimension. For the ENZYMES dataset, higher classification accuracy can be obtained when the feature dimension of the hidden layer nodes in GCN is 32 or 64.

For GFS-GAT, Fig. S3(b) of the *Supplementary Materials* section shows that the classification accuracy decreases slowly with the increasing feature dimension of the hidden layer nodes for the five datasets AIDS, BZR, COX2, PROTEINS, and DHFR, indicating again that better classification performance can be obtained with a lower dimension. For PROTEINS_full dataset, the classification accuracy increases slightly with feature dimension. For ENZYMES and Cuneiform datasets, higher classification accuracy can be obtained when the feature dimension is 32 or 64.

For GFS-GraphSAGE, Fig. S3(c) of the *Supplementary Materials* section shows that the classification accuracy of slowly decreases with the increasing feature dimension for the five datasets AIDS, BZR, COX2, PROTEINS and DHFR. It also indicates that better classification accuracy can be achieved with a lower feature dimension. For PROTEINS_full, ENZYMES and Cuneiform, the classification accuracy increases with the feature dimension of the hidden layer nodes, and a higher classification accuracy can be obtained when the feature dimension is 128.

Hence, the three GFS models can achieve better classification accuracy for most datasets by setting the feature dimension of nodes to a smaller value. The performance fluctuations are not significant for the different optimal feature dimensions identified for the different datasets. Considering both classification accuracy and computational complexity, a reasonable choice of the feature dimension of GNN hidden layer nodes in the GCPU is in the set of {32, 64, 128} for the GFS consequents.

*3) Regularization Parameter*

This subsection analyzes the influence of the regularization parameter $\alpha$ in the GFS objective function $\mathcal{L}_{GFS}$ on the classification performance, where graph classification of the three datasets ENZYMES, PROTEINS and Cuneiform are taken as examples. Fig. S4(a)-(c) of the *Supplementary Materials* section show the variation in classification accuracy of GFS-GCN, GFS-GAT and GFS-GraphSAGE, respectively, with $\alpha$. When $\alpha$ is within the range of 1e-10 to 1e6, all the three models exhibit better classification performance, suggesting an appropriate value of $\alpha$ can be selected in such a range.

### E. Convergence and Stability Analysis

The convergence of GFS-GCN, GFS-GAT and GFS-GraphSAGE is investigated experimentally, with the COX2 dataset taken as an example. Fig. S5 (a)-(c) of the *Supplementary Materials* section show the training set classification performance over iteration epochs respectively, whereas Fig. S6 (a)-(c) of the *Supplementary Materials* section are the corresponding objective function convergence curves. We can see that for the COX2 dataset, the three GFS models can converge after 50 iterations. Similar results are obtained for the other datasets. The results indicate that the proposed GFS models have good stability and convergence.

## V. CONCLUSION

Traditional fuzzy systems have achieved widespread success due to good data-driven learnability and human-like fuzzy inference ability. However, the systems cannot model graph data scenes efficiently. This paper proposes a novel Graph Fuzzy System (GFS) for graph data modeling by extending the fuzzy systems to the field of graph data learning. We define the general concepts of GFS, including graph fuzzy rule base, graph fuzzy sets and graph consequent processing unit (GCPU). In model and algorithm development, we analyze the main GFS components and propose a specific construction method for GFS. The method uses prototype graphs and graph similarity measures to represent graph fuzzy sets, and kernel *K*-prototype graph clustering (K2PGC) to generate GFS antecedents. Meanwhile, the method utilizes mainstream graph neural networks (GNNs) as the central components in GFS rule consequents, and finally obtains the parameters of the graph fuzzy rule consequents by optimizing the given objective function. To evaluate the performance of the proposed GFS, we conduct experimental studies based on several different benchmark datasets for graph classification task. The results indicate that the proposed GFS performs better than traditional fuzzy system modeling methods and mainstream GNNs methods.

Future work on graph fuzzy system modeling will be conducted along three directions. First, we will explore efficient graph clustering algorithms suitable to further improve the robustness of GFS antecedent generation. Second, we will explore possible graph processing operators as the central feature learning module of GCPU in GFS, so as to further improve the flexibility and adaptability of the model. Third, since the GFS discussed in this paper is mainly designed for graph classification, it is not suitable for graph regression, node classification, or link prediction in graph data modeling. Future work will be conducted to further investigate GFS for these three tasks.